\documentclass[conference]{IEEEtran}
\ifCLASSINFOpdf
\else
\fi
\usepackage{amsfonts}
\usepackage{booktabs}
\usepackage{siunitx}
\usepackage{graphicx}
\usepackage{threeparttable, tablefootnote}
\usepackage{url}
\usepackage{cite}

\begin{document}

\title{A Data Driven Approach for Compound Figure Separation Using Convolutional Neural Networks}

\author{
\IEEEauthorblockN{Satoshi Tsutsui \hspace{1in} David J. Crandall}
\IEEEauthorblockA{
School of Informatics and Computing \\ Indiana University, Bloomington, Indiana, USA \\
Email: \{stsutsui, djcran\}@indiana.edu }
}

\maketitle

\begin{abstract}
A key problem in automatic analysis and understanding of scientific
papers is to extract semantic information from non-textual paper
components like figures, diagrams, tables, etc.  Much of this work
requires a very first preprocessing step: decomposing compound
multi-part figures into individual subfigures. Previous work in
compound figure separation has been based on manually designed
features and separation rules, which often fail for less common figure
types and layouts. Moreover, few implementations for compound figure
decomposition are publicly available.
This paper proposes a data driven
approach to separate compound figures using modern deep Convolutional Neural
Networks (CNNs) to train the separator in an end-to-end manner. CNNs
eliminate the need for manually designing features and separation
rules, but require a large amount of annotated training data. We
overcome this challenge using transfer learning as well as
automatically synthesizing training exemplars. We evaluate 
our technique on the ImageCLEF Medical dataset, 
achieving 85.9\% accuracy and 
outperforming previous techniques. We have released our
implementation as an easy-to-use Python library, aiming
to promote further research in scientific figure mining.

\end{abstract}

\IEEEpeerreviewmaketitle

\section{Introduction}

Given the unrelenting pace of science and scientific publication, simply
keeping up with the work in a highly active field can be a daunting challenge.
Researchers increasingly rely on automated techniques to organize,
browse, and search through scientific publications. While modern information
retrieval algorithms can be very successful at analyzing the textual content
of papers, making sense of the other less-structured components of the literature remains a challenge.

For example, scientific papers include a variety of figures, diagrams, tables,
photographs, plots, and other less structured  elements. 
These elements are often crucial to understanding the meaning and potential impact of 
a paper: 
a recent study
discovered a significant correlation between 
properties of a paper's figures and its scientific impact (citation count) in a 
large-scale dataset of biomedical literature~\cite{viziometrics}, for instance.
A variety of specific tasks within this general problem area have been 
studied, including
chart understanding~\cite{Huang2005AssociatingTA}, figure classification~\cite{Cheng2013GraphicalFC,Futrelle2003ExtractionLA}, graphical information extraction~\cite{Huang2007ExtractionOV,Lu2007AutomaticEO}, and pseudo-code detection~\cite{Tuarob2013AutomaticDO}. 

Many figures in scientific papers (over
30\%~\cite{viziometrics,GSB2016}) consist of multiple subfigures, and
so an important preliminary step is to segment or partition them into
their individual components.  Most existing work on figure separation
relies on manually defined rules and features
\cite{lee2015dismantling,Taschwer2016,li5udel}. These techniques
are typically successful for the particular types of figures for which they were designed, but
suffer from the classic problems with rule-based approaches: they tend
to be ``brittle'' because when rules are not satisfied for any given
figure instance, the system fails. For example, an intuitively
reasonable rule is to assume some minimum white space between figures,
but a small percentage of compound figures do not satisfy this
assumption and thus cause the segmentation algorithm to fail, which
likely prevents all subsequent figure understanding steps from
succeeding as well.  This brittleness has driven the document
recognition community, and the entire computer vision and pattern
recognition communities in general, towards more data-driven
approaches that are better able to handle the outliers and
uncertainties that are inherent in visual data.

\begin{figure}[t!]
  \centering
  \includegraphics[width =3.4in]{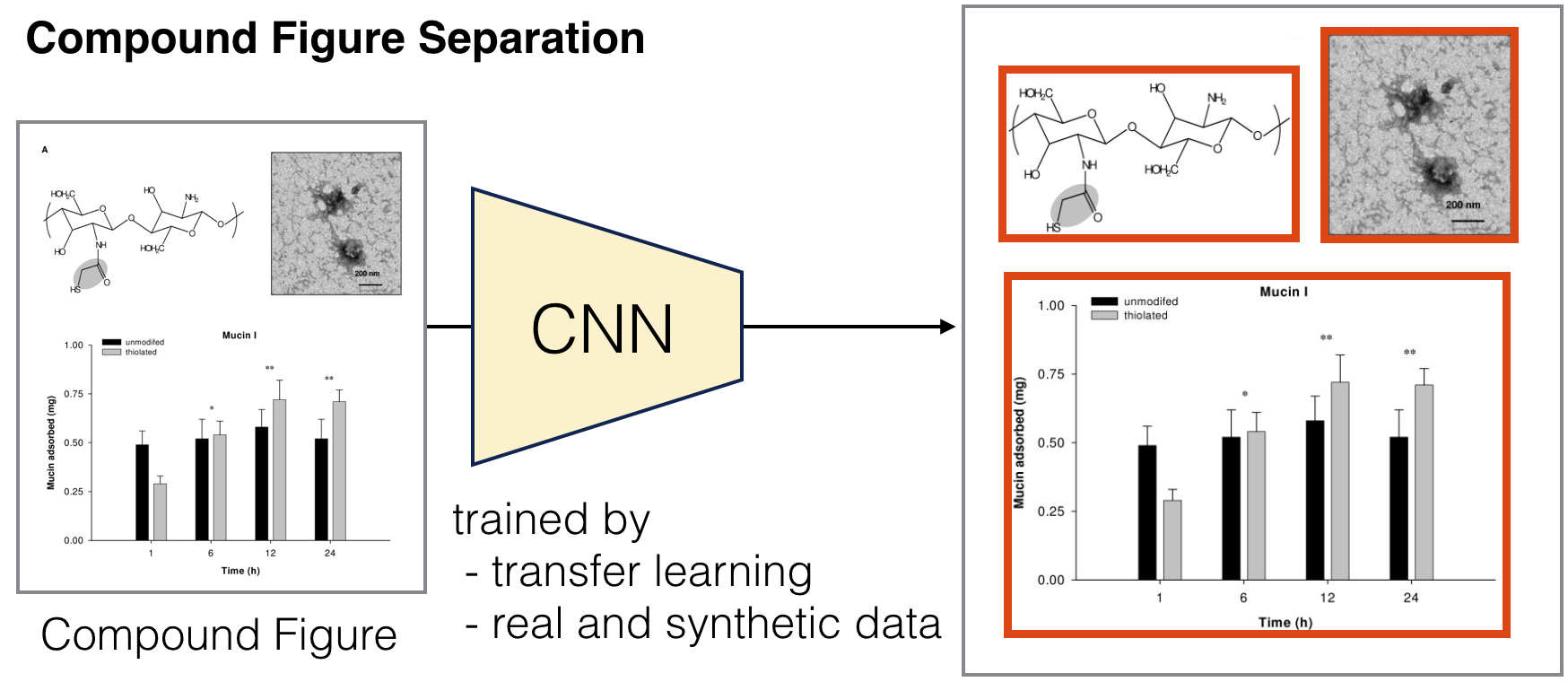}
 	\caption{We propose a new approach for
          segmenting a compound figure into its component
          subfigures. We take a data driven approach using CNNs, with
          transfer learning and exemplar synthesis to overcome limited training datasets.}
	\label{fig:overview}
\end{figure}

In this paper, we propose a data-driven approach to compound figure
separation, eliminating the need for manually designed rules and
features, by using state-of-the-art object detection methods based on
Convolutional Neural Networks (CNNs) \cite{YOLO2}. This approach views the
compound figure separation problem as a form of object detection to
predict bounding boxes of subfigures \cite{GSB2016}, as opposed to
locating the partition boundaries explicitly. However, CNNs require
large amounts of annotated data, a challenge which we address in two
ways. First, we use transfer learning to initialize our CNN with
parameters that were trained on 1.2 million annotated natural images
from ImageNet~\cite{ILSVRC15}. While this idea of fine-tuning by
initializing from a pre-trained model has become common practice in 
computer vision~\cite{yosinski2014transferable,Razavian2014CNNFO}, it is
nevertheless surprising that a problem as different as compound figure
partitioning could benefit from transfer learning of parameters from
consumer images like ImageNet. Second, we augment our training
dataset by ``synthesizing'' new compound figure exemplars by pasting
subfigures onto blank images.  We evaluate our approach on the
ImageCLEF compound figure separation dataset \cite{GSB2016}, and
empirically demonstrate its effectiveness over several baseline
systems that use manually designed features and pipelines.
Finally, we have developed and publicly released an easy-to-use version of our 
compound figure segmentation software via a project website,
\textbf{\url{http://vision.soic.indiana.edu/figure-separator/}}.
To our knowledge, few compound figure separation
tools are publicly available, which we view as a key
 bottleneck for advancing research related to figure mining, especially for scientometric
and bibliometric researchers who may lack computer vision expertise.
We hope our software can push additional work in this area.

To summarize, our paper makes the following contributions:
\begin{itemize}
	\setlength{\parskip}{0cm} 
	\setlength{\itemsep}{0cm} 
	\item We propose a data driven, CNN-based approach for compound figure separation, a problem which has traditionally been addressed with manually-designed pipelines;
	\item We empirically demonstrate the effectiveness of this data-driven approach using transfer learning and synthesized compound figures; and
	\item We have developed an easy-to-use, publicly-available compound figure separation tool in order to encourage figure mining research.
\end{itemize}

\section{Related Work}
\subsection{Scientific Figure Mining}
Understanding scientific figures has long been studied in the document
analysis community. Classifying the type of individual figures is a
fundamental problem
\cite{Futrelle2003ExtractionLA,Cheng2013GraphicalFC}, for example,
while other work attempts to understand figures and extract semantic
information from them \cite{Huang2005AssociatingTA,
  Huang2007ExtractionOV,Lu2007AutomaticEO,IAAI1714275,
  Siegel2016,Choudhury2016ScalableAF}. 
Progress in this area has inspired
research into mining information from figures in large-scale collections of
scientific papers.
A scalable framework to extract figures directly from PDFs
has been proposed~\cite{pdffigures2}.
A figure-oriented literature mining system called
Viziometrics~\cite{viziometrics}, for example, discovered
that each scientific discipline has its own patterns in the
usage of figures, suggesting that scientific fields develop
their own ``visual cultures.'' 
However, most of the above work assumes
that figures consist of one single, simple component, preventing
compound figures that consist of multiple components (which make up at least 30\% of figures in the literature~\cite{viziometrics}) from being successfully analyzed.
Compound figures must first be separated into simpler component
pieces, which is a major motivation for our work.

\subsection{Compound Figure Separation}

This compound figure separation problem has been studied in 
several recent papers~\cite{Siegel2016,lee2015dismantling,Taschwer2016,li5udel}.
Lee \emph{et al.}~\cite{lee2015dismantling} and Siegel \emph{et al.}~\cite{Siegel2016} use background color and
layout patterns, for example, while spaces and lines between subfigures 
are used as cues by Taschwer \emph{et al.}~\cite{Taschwer2016}, and Li \emph{et al.}~\cite{li5udel} use
connected component analysis.
No matter which specific features are used, these approaches
are created through careful engineering using manually-designed rules
and human-crafted features.
In contrast, we adapt a completely data-driven approach
that views compound figure segmentation as an object localization problem, and use modern Convolutional Neural Networks (CNNs) to 
estimate bounding boxes around each of a compound figure's component parts.
This approach avoids the need for manually-written rules, and instead
just requires training data with bounding box annotations. 
This is advantageous because it avoids the ``brittleness'' of
manually-designed recognition pipelines, which often make hidden
assumptions that are easily violated in real instances. It also allows
our approach to be easily customized to the ``visual culture'' of figures
within any specific scientific domain simply by re-training the CNN
on new training data, as opposed to having to re-engineer the system by hand.
Finally, this approach raises the possibility of integrated classifiers that
could perform compound figure separation and subfigure classification in
one unified step, given enough annotated training data.

\subsection{Object Detection}
Compound figure separation is essentially a particular instance of object
(i.e. subfigure) localization. 
The state-of-the-art for object localization in computer vision uses
deep Convolutional Neural Networks, and there are two broad types of popular approaches.
The first is to generate many (thousands of) candidate bounding boxes for potential object
instances in an image, and then use a CNN to classify each bounding box
individually~\cite{r-cnn,fast-r-cnn,faster-r-cnn}. 
An alternative approach is to use a CNN to process
a whole image, and predict classes and bounding box locations at the same
time as a regression problem~\cite{YOLO, YOLO2, SSD}.
These approaches are
usually faster than region-based techniques but with a modest
decrease in detection performance.  
In this paper, we adapt this latter approach, and specifically YOLOv2 \cite{YOLO2}, because it is reported to
be among the fastest and highest performing.

\section{Compound figure separation}\label{sec:method}

We now describe our approach for compound figure separation. 
The heart of our approach is to view figure separation as an
object localization problem, where the goal is to estimate
bounding boxes for each of the subfigures of a compound
figure, using Convolutional Neural Networks. After explaining
the basic model, we discuss how to address the practical (but critical)
problem of training a CNN for our problem given a limited
quantity of training data. 

\subsection{Convolutional Neural Network}

Our general approach is to apply 
the You Only Look Once version 2 (YOLOv2) system
\cite{YOLO2} to our problem of subfigure detection, 
because it is fast, unified, and simple, but highly effective
for object detection.
Please see~\cite{YOLO2} for full details; here we briefly highlight the key properties
of this technique. Unlike prior CNN-based techniques for object localization (e.g.~\cite{r-cnn,fast-r-cnn,faster-r-cnn}),
YOLO avoids the need for separate candidate generation and candidate classification stages, instead
using a single network that takes an image as input and directly predicts bounding box locations as output.
The CNN has 19 convolutional layers, 5 max-pooling layers, and
a skipping connection in a similar manner to residual networks
\cite{he2016deep}. No fully connected layers are included, so the
resolution of the input image is unconstrained.
This makes it possible to train
on randomly resized images, giving the detector's robustness to 
input image resolution. The CNN downsamples the image by a
factor of 32.
Each point
in the final feature map predicts bounding boxes, confidences, and
object classes, assuming that the object is centered at the
corresponding receptive field in the input image.

We follow most of the same implementation settings proposed by the YOLOv2 authors~\cite{YOLO2}. Briefly, we
use stochastic gradient descent to train 160 epochs with learning rate
of 0.001 decreased by a factor of 10 at epochs 60 and 90. We use
a batch size of 64, weight decay of 0.0005, and  momentum of 0.9. For the implementation, we
use Darknet \cite{darknet13}, a
fast and efficient library written in C. After training is done, we
port the trained model into Tensorflow \cite{abadi2016tensorflow} and
use it as the backend of our figure separation tool because it is easier
to customize using Python. Our default input resolution is $544 \times
544$ pixels.

\subsection{Transfer Learning}

Unfortunately, we found that the simple application of YOLOv2 to compound figure
partitioning did not work well. The problem is that
while deep
learning 
with CNNs has had phenomenal success
recently \cite{lecun2015deep}, it requires huge training datasets -- typically hundreds of thousands to millions of images -- which are
not available for many problems. Fortunately, this problem can be at least partially addressed through 
transfer learning \cite{yosinski2014transferable}, where a classifier trained on one problem can be used as a starting
point for a completely different problem. For example, Razavian \textit{et al.}~\cite{Razavian2014CNNFO}
demonstrate that once a CNN is trained to classify natural images
into 1,000 categories using 1.2 million images from ImageNet
\cite{ILSVRC15}, it can be used as a generic feature extractor for
a variety of \textit{unrelated} recognition tasks from natural images. 
Another common trick is to use the CNN parameters trained on ImageNet as initialization for
re-training on a different problem, even in a very different domain
like document images~\cite{harley2015evaluation,deepdocclaccify,Siegel2016},
even though 
ImageNet  does not include any documents. Based on
these reports, we hypothesized that initializing our CNN using ImageNet
pre-trained weights may also be  effective for compound figure separation
problems, and empirically validated this hypothesis.

\subsection{Compound Figure Synthesis}

We found that transfer learning did not completely solve the problem of limited training data, however. 
Although transfer learning helps the CNN initialize its weights to reasonable values, successful training 
still requires that the learning algorithm sees a diverse and realistic sample of compound figures.
To augment the training set, we used our limited ground-truth training set of ``real'' compound figures to automatically generate
a much larger set of synthetic yet realistic additional figures.

\begin{figure}[t!]
  \centering
  \includegraphics[width =2.3in]{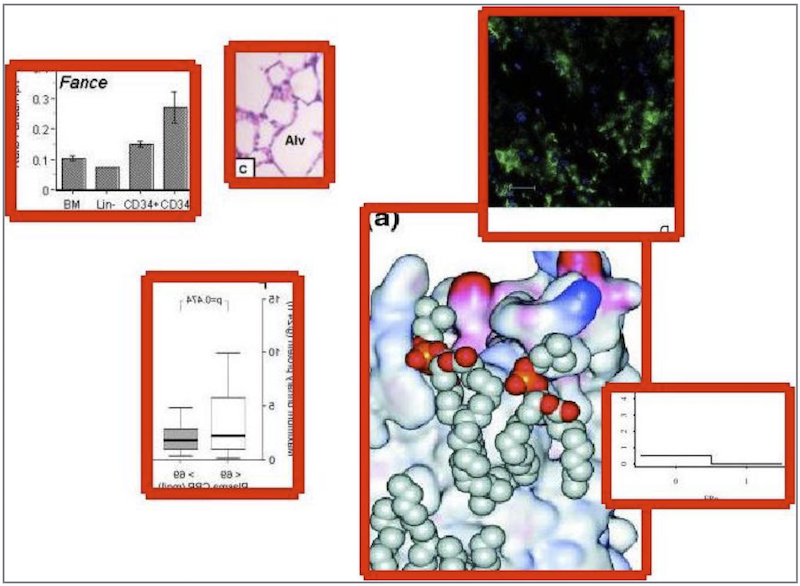}
 	\caption{Sample randomly-synthesized compound figure.  Bounding boxes are displayed for visualization (but of course are not in the actual training data).}
	\label{fig:synthesis_samples_rand}

\end{figure}

We tried two approaches.
The first involves generating compound figures by simply pasting subfigures together in a random manner.
We first create a blank image with a randomly-generated aspect ratio between 0.5 and 2.0. We then choose a subfigure at random from the
``real'' training set, and then rescale the figure with a random scaling factor (while making sure that the subfigure does not exceed the size of
the blank image). Then we randomly select an empty spot within the compound figure (where empty means that the intersection over union (IOU) with existing subfigures is less than 0.05), and
paste the subfigure at this position. If no such empty spot can be found, we end the synthesis. An example of a generated compound figure is shown in Figure
\ref{fig:synthesis_samples_rand}.

Unfortunately, this technique for generating synthetic training examples did not
improve the accuracy of the trained classifier.
We thus tried a second technique that generated compound figures in a more structured way.
Most scientific figures are not composed of randomly-arranged subfigures, of course, but
instead tend to have a more structured layout so that the subfigures are aligned in a near grid-like
pattern. 

\begin{figure}[t!]
  \centering
  \includegraphics[width =2.5in]{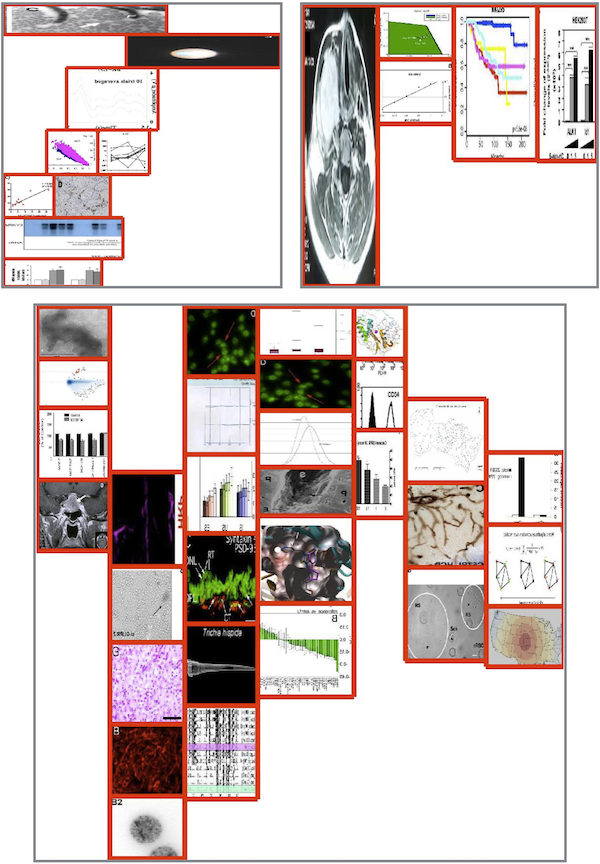}
 	\caption{Three sample synthetic compound figures generated with our grid-based technique. Bounding boxes are displayed for visualization but are not in the actual training data.}
	\label{fig:synthesis_samples}

\end{figure}

We thus first randomly choose a number of rows (between 3 and 7) and a
random height for each row.  Then for each row, we randomly choose a
number of subfigures (between 1 and 7),\footnote{The maximum of 7 is 
an arbitrary choice, although we believe it to be reasonable because
the maximum number of subfigures in our ground truth set of compound figures was about 40.} and paste that number of randomly-selected
subfigures (re-sized to fit the row height) in the row.  To make the
exemplars as difficult as possible, we do not add white space between
the subfigures.  We then randomly transpose rows and columns to
generate additional multiple exemplars for each synthetic compound
figure. Three examples of synthesized figures are shown in Figure
\ref{fig:synthesis_samples}.
Finally, we add additional diversity with three additional manipulations: synthetic images are randomly inverted so that some have black backgrounds instead of white, we randomly apply color transformations to figures to create diversity in color, and we
randomly flip figures horizontally to add diversity in the spatial dimension.

As we show in the next section, this technique for synthesizing
training images significantly improved the performance of the trained detector.
Of course, this is just one technique for generating synthetic
compound figures, and we do not claim it to be the best.

Synthesizing compound figures may at first seem easy, but is actually
difficult if we want to obtain ones similar to real figures without
injecting harmful biases into the training set.
Investigating how to better synthesize real figures is an interesting
direction for future work.

\section{Experiments}

We used the ImageCLEF Medical dataset \cite{GSB2016}, which (to our knowledge)
is one of the largest available collections of figures with bounding box annotations.
The dataset has two versions: the 2015 version has 3,403 training images and 3,381 test images,
while the 2016 version is larger, with 
 6,783
training images and 1,614 test images.
It is reported that the 2015 data is much easier than 2016
data \cite{GSB2016}, so we focus on the 2016 data except when comparing
to baselines for which only 2015 results are available.

We evaluate accuracy using the same metrics defined by the ImageCLEF
task \cite{GSB2016}.  We briefly summarize the metrics here; please
see \cite{GKD2013,Taschwer2016} for full details. For each compound
figure, an accuracy ranging from 0 to 1 is defined as the number of
correctly detected subfigures over the maximum of the number of
ground-truth subfigures and the number of detected subfigures.  A
subfigure is considered to be correct if the area of overlap between the ground
truth and detected boxes is greater than 0.66. Note that this scoring function
penalizes not only missed or spuriously-detected subfigures, but
also multiply-detected subfigures.
The accuracy for a whole dataset is
the average of the individual accuracies.
We also evaluate using mean average precision (mAP)~\cite{Everingham10},
which is a standard measure used in object detection
 and roughly corresponds to the area under the
precision-recall curve.

\begin{table}[!t]
\centering
 \caption{Performance comparison}
\begin{tabular}{@{}l@{\,\,\,\,\,}c@{\,\,\,\,\,}c@{\,\,\,\,\,}c@{\,\,\,\,\,}c@{\,\,\,\,\,}c@{}}
\textbf{Method} & \textbf{Dataset} & \textbf{Accuracy} & \textbf{Precision} &  \textbf{Recall} & \textbf{mAP}  \\
\hline
Lee \textit{et al.}~\cite{lee2015dismantling} & 2016 & 0.566 & 0.824 & 0.378  & --- \\
Li \textit{et al.}~\cite{li5udel} & 2016 & 0.844  & / & / & --- \\
Pure data & 2016 & 0.833  & 0.881 & 0.709 & 0.698  \\
Transfer & 2016 & 0.846 & 0.875 & 0.751 & 0.773  \\
Transfer + random syn & 2016 & 0.842  & 0.873 & 0.726 & 0.746  \\
Transfer + grid syn & 2016 & \textbf{0.859} & \textbf{0.880} & \textbf{0.775} & \textbf{0.782} \\
\hline
Taschwer \textit{et al.}~\cite{Taschwer2016} & 2015 & 0.849  & / & / & --- \\
Transfer + grid syn & 2015 & \textbf{0.917} & \textbf{0.918} & \textbf{0.896} & \textbf{0.889}  \\
\hline
\end{tabular}
\begin{tablenotes}\footnotesize
\item --- indicates an inapplicable metric because the method does not produce confidence values. 
\item ~/~ indicates a value not reported in the original paper and that no public implementation was available
for us to compute it. 
\end{tablenotes}
\label{tbl:results}

\end{table}

Results of our evaluation are shown in Table~\ref{tbl:results} for
several different variants of our detector: \textbf{Pure data} was
trained just on the ``real'' figures in the ground truth, \textbf{Transfer}
was pre-trained using ImageNet and then trained on the ground truth figures,
\textbf{Transfer + random syn} used transfer learning but also synthetic images
using the first of the two techniques described in Section III, 
and \textbf{Transfer + grid syn} used transfer learning and augmented the training
set with the grid-structured synthetic images. For the synthetic settings, we added
a number of synthetic images equal to the number of ground truth images (i.e. we doubled
the training set size).

\begin{figure}[!t]
  \centering
    \includegraphics[width=3in]{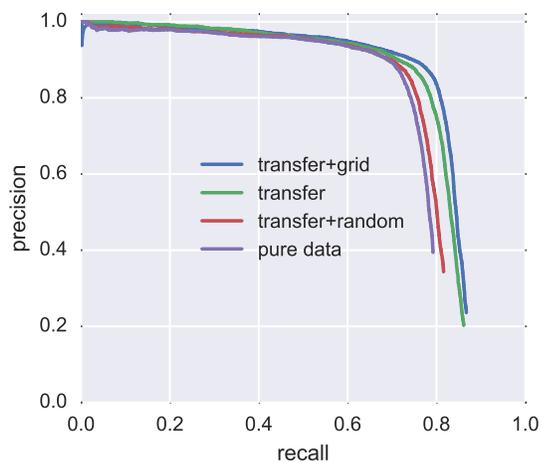}
    \caption{Precision-recall curve of our approach for ImageCLEF Medical 2016 Compound Figure Separation Dataset \cite{GSB2016}}
    \label{fig:pr-curve}
\end{figure}

We observe that pre-training yields a significant improvement, increasing mAP from 0.698 to 0.773.
Adding random synthetic data slightly harms performance, but the grid-structured synthetic data yields a small additional improvement (0.782 vs 0.773).
The 
precision-recall curves in Figure \ref{fig:pr-curve} show a similar story.

The table also compares with three previous algorithms as baselines:
Lee \textit{et
  al.}~\cite{lee2015dismantling}, which uses background color and
layout patterns (and was used in a project mining millions of figures
\cite{viziometrics}), Li \textit{et al.}~\cite{li5udel}, which uses
connected component analysis and is reported to be the best-performing
technique on the 2016 dataset, and Taschwer \textit{et al.} \cite{Taschwer2016}, which
cues on spaces between subfigures and line detection and is reported
to give the highest accuracy on the 2015 dataset. The results show
that our technique significantly outperforms all baselines according to all metrics.

\begin{figure*}[th!]
  \centering
{    \includegraphics[width=6in]{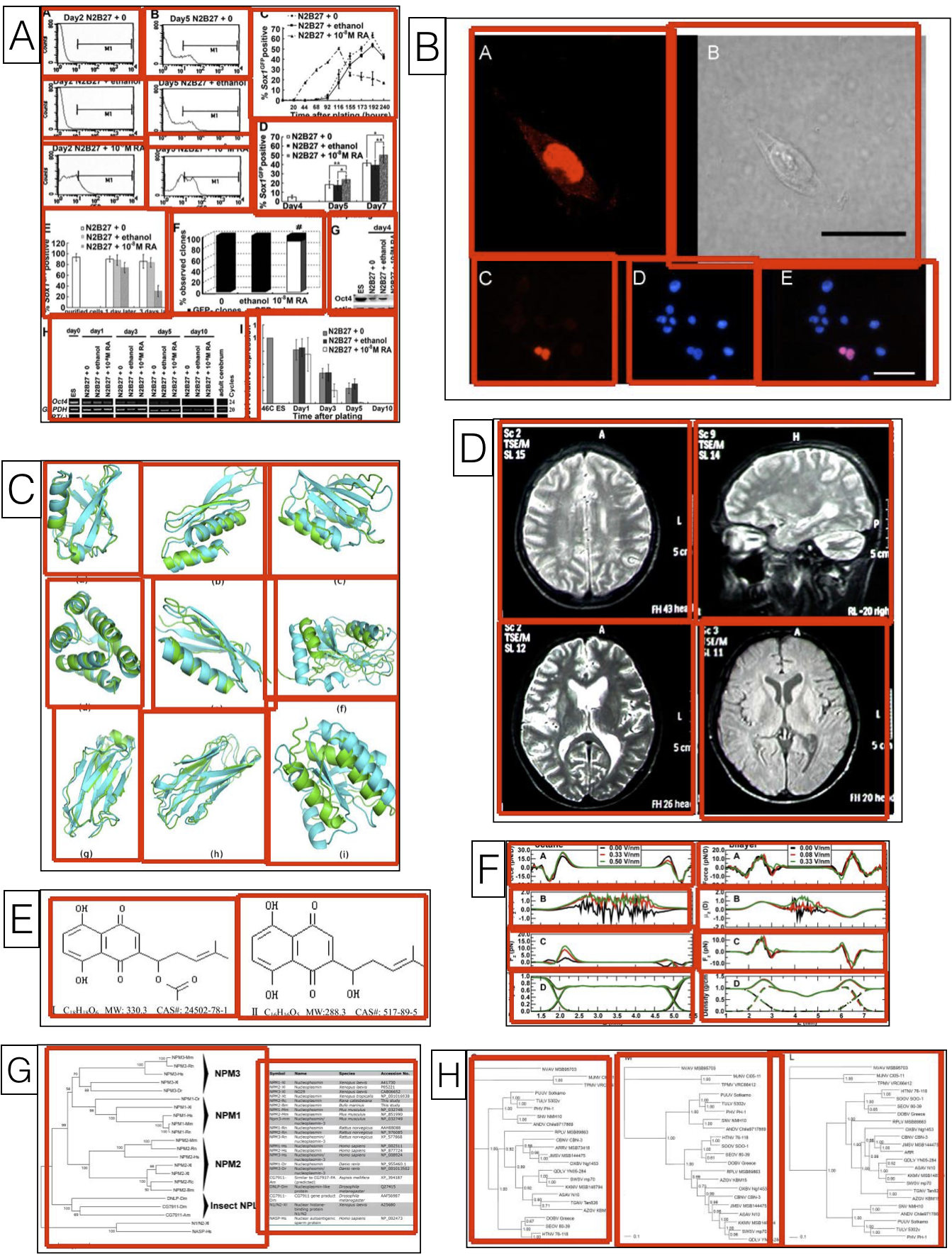}}
    \caption{Correctly separated compound figures. The red bounding boxes show the subfigures extracted by our technique.}

    \label{fig:correct}
\end{figure*}

\begin{figure*}[th!]
  \centering
    \includegraphics[width=6in]{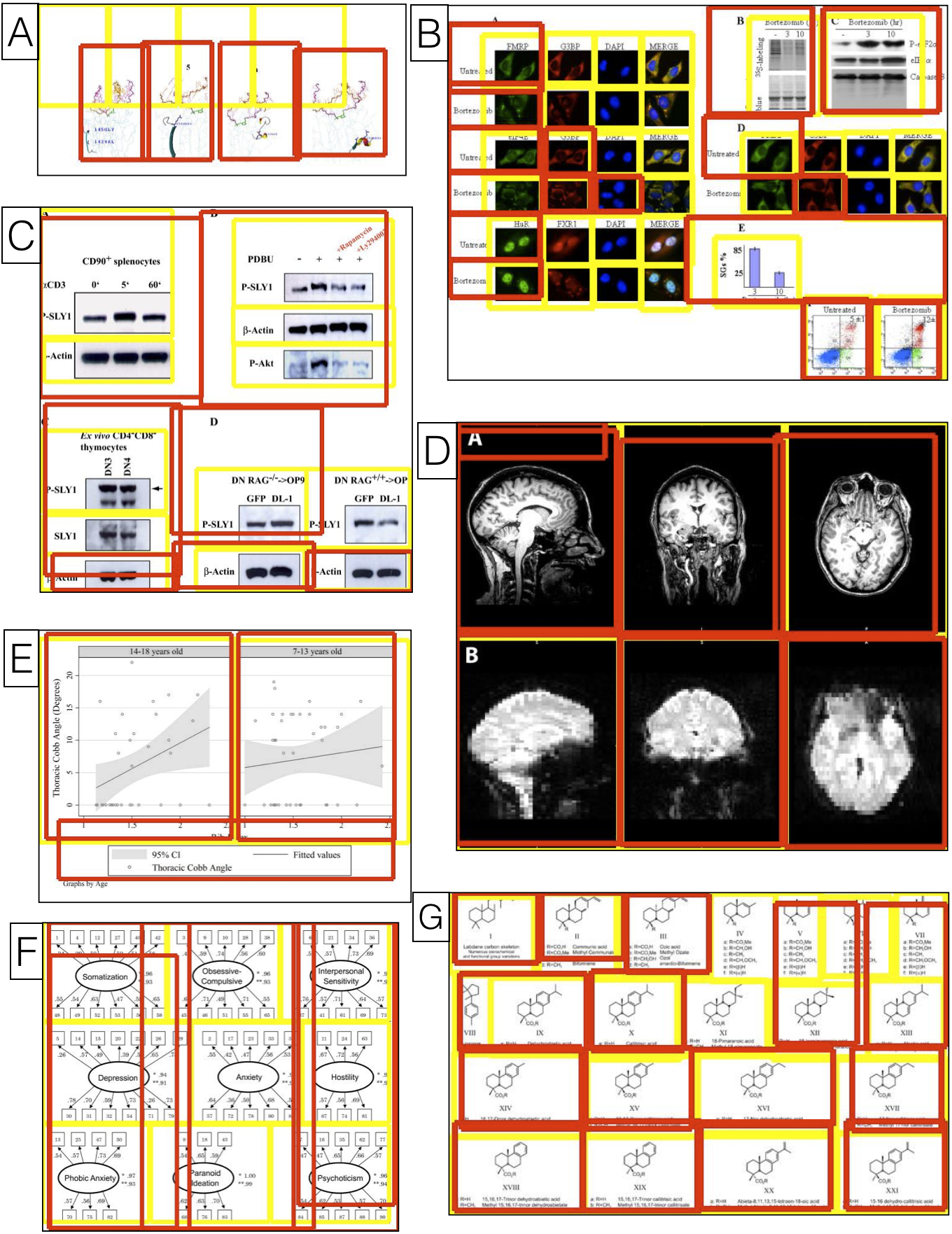}
    \caption{Incorrectly separated compound figures, with the subfigures extracted by our 
technique shown in red and the ground truth shown in yellow.}

    \label{fig:incorrect}
\end{figure*}

Some randomly-sampled correct and incorrect results from our approach are shown in 
Figures~\ref{fig:correct} and~\ref{fig:incorrect}, respectively, where
red boxes indicate our detection results and the yellow boxes are the ground truth.
As seen in Figure~\ref{fig:correct},
our approach is robust
to variations in background color and spaces between subfigures; for example, in panes A, B, D
and F, subfigures are aligned with
almost no spaces and sometimes with black backgrounds. 
Note the wide variety of different figure and subfigure types, layouts, and designs that 
our detector is able to handle.
On the other hand, some kinds of compound figures do confuse our approach,
especially when small and similar subfigures are aligned very closely.
Examples include pane B of Figure \ref{fig:incorrect}, where many
small black images appears in a grid, and G
where many similar chemical compounds are
densely aligned. Other errors happens when relatively large
sub-components are split apart into multiple subfigures, such as the 
legend in pane E, or the ``A'' label in pane D. In many of these cases,
the definition of what should constitute a component is ambiguous, 
with inconsistencies in the ground truth itself.
For instance, the ground truth separates the results of a chemical
experiment  in Figure \ref{fig:incorrect}C, but
not in a similar subfigure in the upper right of Figure \ref{fig:incorrect}B.
We also note that ground truth annotations are
occasionally incorrect such as in Figure \ref{fig:incorrect}A, 
where our algorithm produced much more reasonable results than the
actual ground truth (but was thus penalized in the quantitative evaluation).

\section{Discussion}\label{sec:discussion}
We now discuss important points about our approach and experiments, as well as interesting directions for future work.

\subsection{Compound Figure Synthesis}\label{sec:discussion-syn}

Our approach for synthesizing compound figures is simple and  ad-hoc,
and other approaches are certainly possible.
It turned out to be surprisingly difficult to synthesize realistic
compound figures; 
we observed that unrealistic synthetic training images (such as our first technique of randomly pasting subfigures)
can actually confuse the training algorithm more than they help.

Future
work could explore learned, data-driven approaches to figure synthesis,
such as modeling the probability distribution of figure layout conditioned
on neighboring figures, or on the type or subject of the publication.
Once the parameters of such a generative model are estimated, 
we could sample from that distribution to 
synthesize layouts similar to the
real ones. 

\subsection{Figure Separation Tool}
To our knowledge, implementations of only two figure mining tools
\cite{pdffigures2,Siegel2016} are public, and no compound figure
separation tool is available. 
In fact, one of the major difficulties we experienced in trying to compare
our technique to existing baselines was this lack of publicly-available
implementations.
More importantly, this makes 
it difficult for scientometric researchers, who usually lack 
computer vision expertise, to work on figure mining research. 
This may explain why much work on
mining scientific papers has 
ignored visual aspects \cite{viziometrics} and mainly
focused on data that is easier to use, such as citations
\cite{zhao08citation}, authors \cite{ding11coauthor}, and textual
content \cite{ding2014content}. 
We believe that this is 
because of the much greater availability of  tools for
analyzing textual content in the form of easy-to-use natural language processing libraries~\cite{song2014text}.

To help correct this limited availability of practical tools, we have
made our compound figure separation code publicly available at \url{http://vision.soic.indiana.edu/figure-separator/} as an easy-to-use Python library.  We hope
this may help to foster figure mining research even outside the
computer vision and document analysis communities.  Of course, we are
aware that separating compound figures is not sufficient to apply
large scale figure mining, and that we also need other components such
as figure type classifiers. Developing and releasing implementations
of these components is important future work.  In fact, CNN-based
object detectors may be capable of performing 
figure separation and classification simultaneously.

\subsection{Limited data}

Recent advances in computer vision are due, to a large extent, to
the growing size of annotated training data; ImageNet, for example, has many millions of labeled images.
We believe that the lack of annotated data is holding back scientific figure mining research.
For example, the ImageCLEF Medical dataset
\cite{GSB2016}, the largest available dataset for compound figure
separation, has only 7,000 images for training, which is smaller than
most modern object detection datasets such as PASCAL VOC \cite{Everingham10} or
MSCOCO \cite{lin2014microsoft}. ImageCLEF ground truth also has some 
erroneous annotations for subfigure locations \cite{Taschwer2016}; we found at least 10 erroneous annotations when performing
our experiments (e.g. Figure \ref{fig:incorrect}A). Moreover, the ImageCLEF data only has bounding boxes
and does not have subfigure type annotations.

An important goal for this community could be to build up a much larger
size dataset, perhaps on the order of 100,000 scientific figures with semantic annotations, in order to further accelerate progress in this domain.
Our tool could ease the manual labor involved in creating such a dataset by generating initial subfigure separation proposals which could then
be corrected or refined by a human annotator.

\subsection{Speed}

Many existing techniques first classify figures into compound or
simple, and then run the compound figure separation algorithm only on
the compound figures, in part because the separation algorithm is
relatively slow~\cite{Taschwer2016,viziometrics}.  For example,
separation is reported to take 0.3 seconds per compound figure in
Taschwer \textit{et al.} \cite{Taschwer2016}.  This two step approach
can be dangerous because if a compound figure is not recognized
correctly, the subfigures can never be extracted. In contrast, our
CNN-based separation tool requires 0.12 seconds per figure on a single 
NVIDIA Titan X GPU, which we believe is fast enough to eliminate the need for
compound figure recognition. This takes approximately 33 hours to separate a million compound
figures, and could be trivially parallelized on multiple GPUs.

\section{Conclusion}
We introduced a data driven approach for compound figure
separation, which in the past had been addressed by manually-designed
features and rules. Modern machine learning, and in particular Convolutional Neural Networks, eliminate the need for manual engineering
but require large annotated training datasets. We addressed this challenge through a combination of
transfer learning and automatic synthesis of training exemplars: transfer
learning takes advantage of the visual patterns learned from 1.2 million annotated images from
ImageNet, while compound figure synthesis offers a way to increase the training
data without additional annotation costs. Our experiments demonstrate that our
approaches are effective and outperform previous work.

 We also released an easy-to-use
compound figure separation tool, and hope the tool will help push forward research into scientific figure mining.

\section*{Acknowledgments}
We thank the authors of Lee \textit{et al.}~\cite{lee2015dismantling}
for providing an implementation of their work, Eriya Terada for
providing useful comments on our manuscript, Prof. Ying Ding for her support and  advice,  and the anonymous reviewers for their time and helpful feedback. 
Satoshi Tsutsui is supported by the Yoshida Scholarship Foundation in Japan. 
This work was also supported in
part by the National Science Foundation (CAREER IIS-1253549) and
NVidia, and used the Romeo FutureSystems Deep Learning facility,
supported by Indiana University and NSF RaPyDLI grant 1439007.

\bibliographystyle{IEEEtran}

\bibliography{references}

\end{document}